# SCALE-CONSTRAINED UNSUPERVISED EVALUATION METHOD FOR MULTI-SCALE IMAGE SEGMENTATION

*Yuhang Lu, Youchuan Wan, Gang Li*

School of Remote Sensing and Information Engineering, Wuhan University, P.R. China

**ABSTRACT**

Unsupervised evaluation of segmentation quality is a crucial step in image segmentation applications. Previous unsupervised evaluation methods usually lacked the adaptability to multi-scale segmentation. A scale-constrained evaluation method that evaluates segmentation quality according to the specified target scale is proposed in this paper. First, regional saliency and merging cost are employed to describe intra-region homogeneity and inter-region heterogeneity, respectively. Subsequently, both of them are standardized into equivalent spectral distances of a predefined region. Finally, by analyzing the relationship between image characteristics and segmentation quality, we establish the evaluation model. Experimental results show that the proposed method outperforms four commonly used unsupervised methods in multi-scale evaluation tasks.

## 1. INTRODUCTION

The purpose of image segmentation evaluation is to measure the approaching degree of a segmentation to the human interpretation of an image. Evaluation methods are divided into two categories according to whether user assistance is needed: supervised evaluation and unsupervised evaluation [1]. Supervised methods evaluate the segmentation by comparing it with a manually segmented ground truth image. Such methods are subjective and time consuming, but their results are relatively reliable [2]. They are suitable benchmarks against which different segmentation algorithms can be compared. By contrast, unsupervised methods evaluate the segmentation directly by its own statistical characteristics. Thus, unsupervised evaluation is a more objective and accessible way [3]. Furthermore, it allows for real-time comparison of segmentations and further enables self-tuning of segmentation algorithm parameters [1]. In this paper, we mainly discuss unsupervised evaluation methods.

To evaluate segmentations objectively, we need to answer "what constitutes a good segmentation". For this question, Haralick and Shapiro provided a widely-accepted answer [4], which can be briefly summarized as, a good segmentation should simultaneously have high intra-region homogeneity and inter-region heterogeneity. Most existing unsupervised methods are established based on this criterion. Their general approach is to select certain characteristics to describe homogeneity and heterogeneity quantitatively, and then aggregate these characteristics into an overall goodness score. The commonly used characteristics for intra-region homogeneity include spectral variance, spectral standard deviation, and region entropy. The characteristics for inter-region heterogeneity mainly include spectral distance, spatial autocorrelation and layout entropy.

Previous unsupervised methods share the common problem of not considering the application scenario when evaluating segmentations. In other words, their evaluation results for a segmentation under different conditions are all the same. Actually, such an outcome is unreasonable for most natural images, especially for remote sensing images, because the user's requirement for the segmentation result is changed with application scenarios. In practical applications, this requirement always refers to the segmentation scale. For example, when segmenting urban remote sensing images, if a detailed segmentation is required, cars may have to be isolated from the road, whereas if a coarse segmentation is required, cars should be put in the same region with the road. Thus, segmentation quality is related to the target scale. To this end, we propose a scale-constrained evaluation method that can evaluate the relative quality of a segmentation.

The rest of this paper is organized as follows. Section 2 briefly reviews the existing unsupervised evaluation methods. Section 3 presents the proposed scale-constrained evaluation method. Section 4 validates the effectiveness of the proposed method by experiments. Finally the conclusion is drawn in Section 5.

## 2. RELATED WORK

Early studies in unsupervised evaluation are empirically based. Their evaluation functions rely on empirical parameters. For example, the metric $F$ proposed by Liu and Yang [5] measures segmentation quality with the use of intra-region spectral standard deviation and the number of regions. On the basis of their work, Borsotti et al. [6] proposed an

Supported by the National Science and Technology Support Program (No. 2014BAL05B07), and the National High-tech R&D Program of China (No. 2013AA122104).

improved metric $Q$ to decrease the bias toward both over-segmentation and under-segmentation. However, the application range of their methods is limited by the empirical parameters.

Subsequently, some rule-based methods were proposed. They typically used certain characteristics to instantiate intra-region homogeneity and inter-region heterogeneity and yield a goodness score, such as the metric $F_{RC}$ proposed in [7] and the $E$ in [2]. The former used intra-region spectral variance and inter-region spectral distance, whereas the latter used region entropy and layout entropy. In addition, other metrics of this type, such as $Zeb$ [8] and $V_{EST}$ [9], are well summarized in [1].

In recent studies, more image characteristics have been explored to describe homogeneity or heterogeneity. In [10], the Markov random field model is introduced to measure the contrast between the interior and the exterior of the edge of a segmentation. In [11], the spatial autocorrelation metric, Moran's I, was employed to measure inter-segment goodness. Corcoran et al. [12] defined the spatial separation and cohesion of segmentations. In [13], the segmentation quality was decomposed into goodness-of-fit energy and complexity energy, and a parameter to control the weight of both energies was set to evaluate segmentations at multiple scales; however, the selection of the parameter is left to the user.

Moreover, with the development of machine learning, some unsupervised evaluation frameworks based on machine learning algorithms have been proposed. For instance, Chabrier et al. [14] used genetic algorithms to combine six different evaluation criteria to obtain an optimized result. Peng and Veksler [15] trained a classifier for evaluation purposes by AdaBoost; the feature space includes intensity, texture, gradient direction, and corners of the segmentation. The work proposed in [16] is designed in a similar way except for the training model, which is a Bayesian network, and the input features, which are four low-level image features.

### 3. SCALE-CONSTRAINED EVALUATION METHOD

In this section, we present the scale-constrained unsupervised evaluation method. It uses regional saliency and merging cost to measure intra-region homogeneity and inter-region heterogeneity at first, and then they are standardized into a kind of intra- and inter-region spectral distance. Finally, the absolute quality for the segmentation's own scale and the relative quality for the target scale are defined. The flowchart is shown in Figure 1.

#### 3.1. Intra-region homogeneity

Saliency detection is widely used to extract regions of interest in images. When the object is change from an image into a single region, the standout pixels within the region can be detected as well. Thus regional saliency could be the measure of intra-region homogeneity. The saliency detection algorithm we adopted in this study is the frequency-tuned (FT) algorithm [17]. Its basic idea is to subject the image to a low-pass filter and to a high-pass filter, and then subtract the two filtering results to obtain the saliency map.

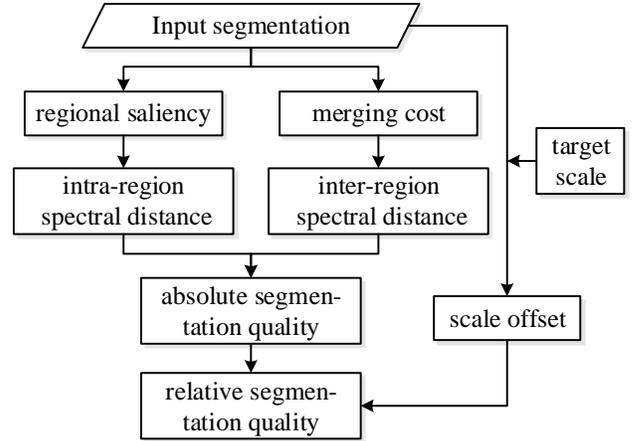

**Fig. 1**. Flowchart of the proposed method

Let $I$ be the original image, $S$ be the segmentation to be evaluated, $w$ be the width of $I$, $h$ be the height, and $N$ be the area (i.e., $N = w \times h$). In $S$, $n$ regions exist, namely, $R_1, R_2, \ldots, R_n$; the area of region $R_i$ is $N_i$.

First, compute the mean pixel value of image $I$ in $Lab$ color space. The result is a $[L, a, b]^T$ vector, which is denoted by $\mathbf{I}_\mu$.

Then filter the image by the $DoG$ filter, which is defined as:

$$DoG(x,y) = \frac{1}{2\pi}\left[\frac{1}{\sigma_1^2}e^{-\frac{(x^2+y^2)}{2\sigma_1^2}} - \frac{1}{\sigma_2^2}e^{-\frac{(x^2+y^2)}{2\sigma_2^2}}\right] \quad (1)$$

In image applications, a Gaussian convolution kernel is often used to express this filter approximately. The kernel used in FT is $\frac{1}{16}[1,4,6,4,1]$. Applying it to convolve with the image, a Gaussian blurred image of $I$, which is denoted by $\mathbf{I}_\omega$, is obtained.

The saliency of pixel $(x, y)$ is formulated as:

$$sal(x, y) = \|\mathbf{I}_\mu - \mathbf{I}_\omega(x, y)\| \quad (2)$$

The overall saliency, $sal_i$, of region $R_i$ is:

$$sal_i = \sum_{(x,y) \in R_i} sal(x, y) \quad (3)$$

#### 3.2. Inter-region heterogeneity

Inter-region heterogeneity refers to the probability of two adjacent regions being over-segmented. If two regions are over-segmented, two possibilities exist: their image features are similar or their areas are small. Thus, the merging cost proposed in [18] is suitable to measure the inter-region heterogeneity.

The merging cost between two regions is given by:

$$cost(i,j) = \frac{N_i \cdot N_j}{N_i + N_j} \cdot \frac{\|\vec{\mu_i} - \vec{\mu_j}\|^2}{c} \quad (4)$$

where $\vec{\mu}$ is the mean spectral vector, and $c$ is the dimension of $\vec{\mu}$.

After all merging costs in the segmentation are computed, the regions are considered the vertices, whereas the merging costs are considered the weights. An undirected region adjacency graph $G = \langle V, E \rangle$, where $V$ is the set of vertices and $E$ is the set of edges, can then be constructed. The global merging cost of the segmentation is defined as the average of all weights as follows:

$$cost = \frac{1}{k}\sum_{i=1}^{k} cost_i \quad (5)$$

where $k$ is the number of edges.

### 3.3. Standardization

The dimensions of regional saliency and merging cost are different; thus, they should be standardized before using them in the evaluation. To be clear, the segmentation's scale is defined as the square root of the average size of the regions. With $s$ denoting the scale of $S$, the following can be obtained:

$$s = \sqrt{N/n} \quad (6)$$

Let $s$ be the target scale, and $sal_i$ be the saliency of region $R_i$. A square region $R'$, whose left half has a gray value of 0, right half has a gray value of $t$, and area equals to $s^2$, is considered. The average pixel saliency of this region is linear with $t$. With the fitting precision $R^2 = 0.9975$, their relationship can be expressed by:

$$sal' = 0.515 \cdot t \cdot s^2 \quad (7)$$

Let $R'$ have the same saliency as $R_i$, then:

$$t = \frac{1.942 \cdot sal_i}{s^2} \quad (8)$$

In equation (8), the regional saliency is transformed into its equivalent intra-region spectral distance.

Similarly, the merging cost can be also standardized into inter-regional spectral distance. Denote $cost_i$ as the merging cost of two regions. Two adjacent square regions are considered, with the left one having a gray value of 0, the right one having a gray value of $t$, and their areas being $s^2$. According to equation (4), the merging cost of them is:

$$cost' = \frac{1}{2} \cdot s^2 \cdot t^2 \quad (9)$$

Let $cost'$ equals to $cost_i$, then:

$$t = \frac{\sqrt{2 \cdot cost_i}}{s} \quad (10)$$

Thus far, the standardized results of both regional saliency and merging cost are obtained. The intra-regional spectral distance of $S$ is denoted by $d_{intra}$, and the inter-regional spectral distance is denoted by $d_{inter}$. Let every region in $S$ have the same weight, then the following can be derived:

$$d_{intra} = \frac{1}{n}\sum_{i=1}^{n} \frac{1.942 \cdot sal_i}{s^2} \quad (11)$$

$$d_{inter} = \frac{1}{k}\sum_{i=1}^{k} \frac{\sqrt{2 \cdot cost_i}}{s} \quad (12)$$

### 3.4. Segmentation quality

The relationship between image characteristics and segmentation quality is always defined empirically in previous methods. In this section, we attempt to determine the regularity between intra- and inter-region spectral distances and segmentation quality.

A gray image in gradient color is considered; the gray value of each column is constant, whereas the gray value of each row is $0, 1, \cdots, 255$ from left to right. No matter how many equal parts an image is split into by column, the absolute quality of the segmentation is assumed to remain constant. For convenience of calculations, we set the image size to $256 \times 256$. Then, we split the image into $n$ equal parts by column; the variations in $d_{intra}$ and $d_{inter}$ of the corresponding segmentation are shown in Table 1.

**Table 1**. Variations in $d_{intra}$ and $d_{inter}$ with $n$

| $n$ | 1 | 2 | … | $k$ | … | 128 | 256 |
|---|---|---|---|---|---|---|---|
| $d_{intra}$ | 128 | 64 | … | $128/k$ | … | 1 | 0 |
| $d_{inter}$ | N/A | 128 | … | $256/k$ | … | 2 | 1 |

Except for $n = 1$ and $n = 256$, $d_{inter}$ is always twice as high as $d_{intra}$. Let their absolute segmentation quality, $Q_0$, be equal to 1, then

$$Q_0 = \frac{1}{2} \cdot \frac{d_{inter}}{d_{intra}} \quad (13)$$

Given this assumption, the quantity relationship of $d_{intra}$ and $d_{inter}$ with the absolute segmentation quality is expressed by equation (13).

Finally, with $s_t$ denoting the target scale, we define the relative segmentation quality as the product of the absolute segmentation quality and the scale offset as follows:

$$Q_t = \frac{d_{inter}}{2d_{intra}} \cdot \frac{\min(s, s_t)}{\max(s, s_t)} \quad (14)$$

## 4. EXPERIMENTAL RESULTS

In this section, we use the supervised evaluation results as the benchmark to compare the proposed method with four commonly used unsupervised evaluation methods. Experiments are performed on the Berkeley Segmentation Data Set (BSDS500), which includes 500 natural images with each image segmented by five different subjects on average [19]. For each image, we select three segmentations at different scale: the largest, the smallest, and a medium one. Figure 2 is taken as the example to illustrate our experiment, where (a) is the original image with a size of $321\times 481$, and (b), (c), and (d) are its ground truths at three different scales.

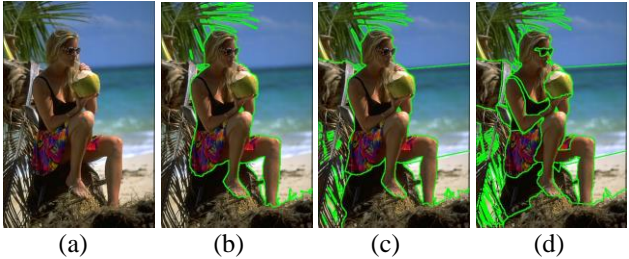

(a) (b) (c) (d)

**Fig. 2.** Example image and three ground truths

The three ground truths are denoted by $S_1$, $S_2$, and $S_3$, and their scales are denoted by $s_1$, $s_2$, $s_3$, respectively. First, the proposed method is used to evaluate the absolute segmentation quality of $S_1$, $S_2$, and $S_3$. The results are shown in Table 2.

**Table 2.** Absolute segmentation qualities of $S_1$, $S_2$ and $S_3$

|  | $S_1$ | $S_2$ | $S_3$ |
|---|---|---|---|
| $n$ | 16 | 58 | 130 |
| $s$ | 98.235 | 51.595 | 34.463 |
| $d_{intra}$ | 65.763 | 56.789 | 51.76 |
| $d_{inter}$ | 43.213 | 46.307 | 41.246 |
| $Q_0$ | 0.329 | 0.408 | 0.398 |

As shown in Table 2, $d_{intra}$ decreases as the scale decreases, indicating that the homogeneity within regions increases normally. By contrast, $d_{inter}$ does not decrease as expected; the $d_{inter}$ value of $S_2$ is even higher than $S_1$, suggesting that some regions in $S_1$ are visually undersegmented. Consequently, the absolute quality of $S_1$ is lower than those of $S_2$ and $S_3$.

Let $s_1$, $s_2$, and $s_3$ be the target scales to evaluate the relative qualities of the segmentations by the proposed method. The result of the proposed method is compared with four commonly used unsupervised methods, namely, $F$, $Q$, $F_{RC}$, and $E$. All the results are detailed in Table 3. The best segmentation in each evaluation is highlighted in bold.

**Table 3.** Unsupervised evaluation results of $S_1$, $S_2$ and $S_3$

|  | $S_1$ | $S_2$ | $S_3$ |
|---|---|---|---|
| $F$ | **0.115** | 0.281 | 0.781 |
| $Q$ | **0.427** | 0.705 | 1.05 |
| $F_{RC}$ | 35.557 | **49.215** | 21.632 |
| $E$ | **2.622** | 2.869 | 3.091 |
| $Q_t(s_t = s_1)$ | **0.329** | 0.214 | 0.14 |
| $Q_t(s_t = s_2)$ | 0.173 | **0.408** | 0.266 |
| $Q_t(s_t = s_3)$ | 0.115 | 0.272 | **0.398** |

Table 3 shows that the proposed method produces different evaluation results according to the target scale. By contrast, the results of other methods are unchangeable. Additionally, except for $F_{RC}$, the other three methods prefer the segmentation at a large scale.

Finally, the supervised evaluation criteria Segmentation Covering is employed to validate the effectiveness of these unsupervised methods [19]. The supervised evaluation result can be represented by a $3\times 3$ matrix, with each row corresponding to a ground truth and each column corresponding to an image. In this experiment, the matrix is
$\begin{pmatrix} 1 & 0.558 & 0.418 \\ 0.516 & 1 & 0.738 \\ 0.327 & 0.648 & 1 \end{pmatrix}$. The correlation coefficients of the evaluation results of $F$, $Q$, $F_{RC}$, $E$, and the proposed method with respect to this matrix are 0.129, 0.164, 0.027, 0.176, and 0.97, respectively.

For all 500 images in BSDS500, the average accuracy values of $F$, $Q$, $F_{RC}$, $E$, and the proposed method are 0.077, 0.047, 0.026, 0.099 and 0.711, respectively. The codes and more experimental results are available at: https://github.com/Rudy423/SegEvaluation.

## 5. CONCLUSION

In this paper, we propose a scale-constrained unsupervised evaluation method for multi-scale image segmentation. The main difference between the proposed method and previous methods is, we believe that the quality of segmentation is dependent on application scenarios. Thus the target scale is introduced into the evaluation model to constrain the relative segmentation quality. Experimental results show that the proposed method performs better than four existing methods in multi-scale evaluation tasks. In our future studies, we will attempt to add more high-level image features, such as textures and semantic information, into the evaluation model.